\documentclass{article}

\usepackage{arxiv}
\usepackage{xcolor}
\usepackage[utf8]{inputenc} % allow utf-8 input
\usepackage[T1]{fontenc}    % use 8-bit T1 fonts
\usepackage{hyperref}       % hyperlinks
\usepackage{url}            % simple URL typesetting
\usepackage{booktabs}       % professional-quality tables
\usepackage{amsfonts}       % blackboard math symbols
\usepackage{nicefrac}       % compact symbols for 1/2, etc.
\usepackage{microtype}      % microtypography
\usepackage{lipsum}		% Can be removed after putting your text content
\usepackage{graphicx}
\usepackage{natbib}
\usepackage{doi}

\title{Segment Anything Model (SAM) Meets Glass: Mirror and Transparent Objects Cannot Be Easily Detected}

\author{Dongsheng Han \\
	Kyung Hee University \\
 \And
Chaoning Zhang\thanks{Correspondence Author: chaoningzhang1990@gmail.com} \\
	Kyung Hee University\\
 \And
Yu Qiao \\
	Kyung Hee University\\
 \And
Maryam Qamar \\
	Kyung Hee University\\
  \And
Yuna Jung \\
	Kyung Hee University\\
\And
SeungKyu Lee \\
	Kyung Hee University\\
 \And
Sung-Ho Bae \\
	Kyung Hee University\\
\And
Choong Seon Hong \\
	Kyung Hee University\\
}

\renewcommand{\headeright}{}
\renewcommand{\undertitle}{}
\renewcommand{\shorttitle}{Segment Anything Model meets Glass} 

\begin{document}
\maketitle

\begin{abstract}
Meta AI Research has recently released SAM (Segment Anything Model) which is trained on a large segmentation dataset of over 1 billion masks. As a foundation model in the field of computer vision, SAM (Segment Anything Model) has gained attention for its impressive performance in generic object segmentation. Despite its strong capability in a wide range of zero-shot transfer tasks, it remains unknown whether SAM can detect things in challenging setups like transparent objects. In this work, we perform an empirical evaluation of two glass-related challenging scenarios: mirror and transparent objects. We found that SAM often fails to detect the glass in both scenarios, which raises concern for deploying the SAM in safety-critical situations that have various forms of glass. 
% In this article, we report the robustness of SAM's pre-trained model on challenging transparent and mirror objects, and investigate its potential for future development in this field. 
\end{abstract}

% keywords can be removed

\section{Introduction}
In the past few years, generative AI~\cite{zhang2023complete} has caught significant attention with interesting applications like ChatGPT~\cite{zhang2023chatgpt}, text-to-image~\cite{zhang2023text}, text-to-speech~\cite{zhang2023audio} and graph generation~\cite{zhang2023graph_survey}. A key factor that drives the development of generative AI is foundation model~\cite{bommasani2021opportunities} that at inference can generalize to tasks and data distributions different from training. With the success of ChatGPT~\cite{zhang2023chatgpt}, GPT-3~\citep{brown2020language} has been widely recognized as one of the most widely recognized foundation models for NLP. 

Very recently, Meta AI research team has recent released a segment anything project~\cite {kirillov2023segment} that introduces a promotable segmentation task for training a vision foundation model. The resulting segment anything model (SAM) has been recognized as the GPT-3 moment for vision. The model was trained on over 1 billion masks on 11 million licensed and privacy-respecting images. 
It represents a significant step towards achieving cognitive recognition for all objects in the world, aiming to handle interactive segmentation tasks while addressing real-world constraints. %Its impressive performance on segmentation has led many researchers to exclaim that the era of computer vision tasks being dominated by visual processing has come to an end.
Segmentation tasks encompass a wide range of challenging areas, and the detection of transparent objects and mirror regions constitutes one of the most representative and challenging tasks in this field. Given the ubiquitous appearance of such glass objects in daily life and the unique challenges in detecting, localizing, and reconstructing them from color images in computer vision\citep{lin2021rich}. This diccicult is due to the fact that most glass objects exhibit a visual appearance that includes both the transmitted background scene and reflected objects\citep{maeno2013light}\citep{he2021enhanced}, while the visual appearance observed from a mirror surface is entirely from reflected objects. These issues led to many incidents, such as collisions of autonomous mobile robots with transparent front doors or mirror walls and robot arms struggling to grip a transparent bottle.

In this work, we examine the performance of the SAM foundation model in recognizing and segmenting transparent objects and mirror surfaces and evaluate its segmentation results. We aim to explore the challenges that arise for foundation models in computer vision tasks when dealing with the reflection and refraction phenomena that occur on glass and mirror objects, as well as the faint object boundaries that result from these effects. After extensive testing on large glass and mirror dataset benchmarks, we find that while SAM segments general objects well in natural images well but often fail to detect mirror and transparent objects. Failing to recognize glass can cause serious issues when deploying the vision foundation model in safety-critical setups. 

% Nonetheless, we believe that the emergence of foundation models like SAM provides a new possibility and challenge for the segmentation of transparent and mirror objects. We hope that this will give confidence to the field of recognition and segmentation of transparent and mirror objects. With further research and development, we believe that the performance of these models can be improved, and that they can contribute significantly to the development of computer vision technologies.

\section{Experimental Evaluations}
\textbf{Datasets:}
As the SAM model is designed to recognize and segment all objects in an image, we select benchmark datasets where only one transparent or mirror object exists in a given scene. This allows us to evaluate the model's capability to identify and segment these specific objects accurately. We conduct experiments on two glass segmentation datasets GDD~\citep{mei2020don}, GSD~\citep{lin2021rich} and two mirror segmentation datasets MSD~\citep{yang2019my}, PMD~\citep{lin2020progressive}. GDD contains 2980 training and 936 test images. GSD is transparent object dataset collected and labeled through networking and photography, consisting of 4102 annotated glass images with close-up, medium, and long shots from diverse scenes. MSD is a large mirror segmentation dataset with 4018 images. 
PMD is the latest mirror segmentation dataset collected from both indoor and outdoor scenes.
% The glass and mirror object categories in the benchmarks mentioned above are relatively consistent across all images, allowing for a somewhat fair comparison. 

In our final additional visualization experiment, we included a multi-class dataset of transparent objects called Trans10k~\citep{xie2020segmenting} in addition to the previous benchmark datasets. Trans10k is a dataset of transparent objects categorized into two groups based on objects properties: movable small objects (things) and non-movable objects (stuff).
 
\textbf{Implementation Details:}
We conduct the test using the SAM best pre-trained model VIT-H, which was pre-trained on a large segmentation dataset (SA-1B) and generously provided as open source by Meta AI Research. As SAM does not specify the object categories for segmentation, it outputs all possible objects in the image. To test SAM's capability to recognize glass and mirror objects, we calculated the Intersection over Union (IoU) between the predicted results and the ground truth for each object outputted by SAM. We then selected the result with the highest IoU with the ground truth for glass objects and used it as the detection result for SAM. We conducted different evaluations based on this result.

\textbf{Evaluation Metrics:}
We employed five commonly used evaluation metrics in semantic and glass surface segmentation tasks: Intersection over Union (IoU), pixel accuracy (ACC), weighted F-measure ($F_{\beta}$) \cite{margolin2014evaluate}, mean absolute error (MAE), and balance error rate (BER).
${F}_{\beta}$ is a harmonic mean of average precision and average recall defined as follows.
% F-measure metric \cite{A bi-directional message passing model for salient object detection} compensates for the deficiencies of Recall and Precision in salient object detection. \cit{Where is my mirror?}, \cite{Don’t hit me! glass detection in real-world scenes}F-measure metric is also used to evaluate the model performance of GDD and MSD datasets.
\begin{equation}
\label{eqn:03}
{\textit{F}_{\beta}}=\frac{(1+{\beta^{2}})(Precision\times{Recall})}{{\beta^{2}}Precision+Recall},
\end{equation}
where $\beta^{2}$ is set to 0.3 as suggested in \cite{achanta2009frequency}.
Mean absolute error (MAE) is widely used in foreground-background segmentation tasks where average pixel-wise error between predicted mask P and ground truth mask G are calculated.
\begin{equation}
\label{eqn:03}
{\textit{MAE}}=\frac{1}{H\times{W}}\sum_{i=1}^{H}\sum_{j=1}^{W}\mid P(i,j)-G(i,j)\mid,
\end{equation}
where P(i, j) indicates predicted probability at location (i, j).
We employ the balance error rate (BER) as an evaluation metric, which takes into account the imbalanced regions in glass and mirror object segmentation tasks. This metric provides a quantitative measure for evaluating the performance of glass surface segmentation.
\begin{equation}
\label{eqn:03}
{\textit{BER}}=(1-\frac{1}{2}(\frac{TP}{\textit{N}_{p}}+\frac{TN}{\textit{N}_{n}}))\times{100},
\end{equation}
where TP, TN, Np, and Nn represent the numbers of true positives, true negatives, glass pixels, and non-glass pixels.
\subsection{Experiment on glass objects}
We compare SAM with the state-of-the-art semantic segmentation, shadow detection, saliency object detection, and glass segmentation methods such as MirrorNet, Translab, GDNet, EBLNet as shown in Table \ref{GDD}, \ref{GSD}. 

As shown in Table \ref{GDD} and \ref{GSD}, We observed that the segmentation results of glass objects in two benchmarks were poor for the SAM model.

\begin{figure*}
\centering
\includegraphics[scale=0.8]{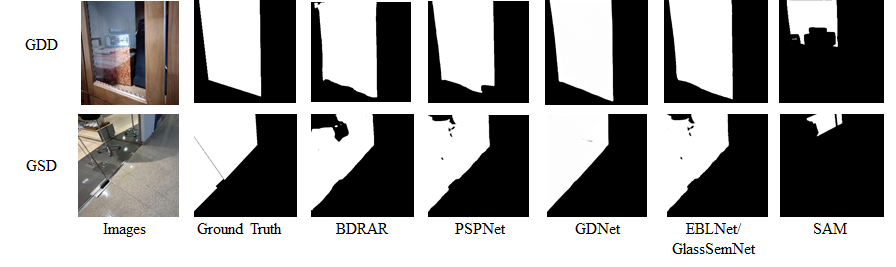}
    %\vspace*{-2mm}
\caption{Sample qualitative comparison results on GDD and GSD.}
  \label{glass} 
\end{figure*}

As shown in Figure \ref{glass} and \ref{glass-all}, We find that the SAM model struggle to recognize and segment the boundary regions, particularly when the boundaries of glass objects are distorte by light. These challenges in glass objects segmentation are caused by the fact that transparent surfaces exhibit a visual appearance that includes both the transmitted background scene and any objects behind them, making it difficult to distinguish and accurately segment them from their surroundings. The phenomenon of transmission tends to cause SAM to detect objects behind glass objects, rather than the glass objects themselves.

% a large amount of data is intentionally captured with oblique angles to enhance the reflection phenomenon in the images. However, we found that the SAM model has limited understanding of reflection and is unable to accurately identify glass objects. 
%GDD
\begin{table}%\scriptsize
    \centering
    \begin{tabular}{lccccc} %全部居中
    \toprule
    Method & IoU $\uparrow$ & Acc $\uparrow$ &${F}_{\beta}$ $\uparrow$ & mAE $\downarrow$ &BER $\downarrow$ \\
    \midrule
    PSPNet \cite{zhao2017pyramid}    &84.06 &0.916 &0.906& 0.084 &8.79\\
    
    PointRend \cite{kirillov2020pointrend}  & 86.51 &0.933&0.928 & 0.067& 6.50\\
    PiCANet \cite{liu2018picanet}   &83.73 &0.916&0.909 & 0.093 &8.26\\
    %\midrule
    DSC \cite{hu2018direction} &83.56 & 0.914  &0.911   & 0.090 & 7.97\\
    BDRAR \cite{zhu2018bidirectional}  &80.01 & 0.902 &0.902    & 0.098 & 9.87\\
    MirrorNet \cite{yang2019my} &85.07 & 0.918 &0.903    & 0.083 & 7.67\\
    GDNet \cite{mei2020don}  &87.63 & 0.939 &0.937 & 0.063 & 5.62\\
    EBLNet \cite{he2021enhanced}  & 88.16 & 0.941  &0.939   & 0.059 & 5.58\\
    %EBLNet(ResNext101) &ICCV’21 & 80.33  & 95.14     & 0.049 &8.63\\
    \midrule
    \textbf{SAM}& \textbf{48.47} &\textbf{0.732} &\textbf{0.798}&\textbf{ 0.268} & \textbf{26.08}\\
    \bottomrule
    \end{tabular}
            %\vspace*{-2mm}
    \caption{Experimental Comparison on GDD}
            %\vspace*{-1mm}
    \label{GDD}
\end{table}
\begin{table}%\scriptsize
    \centering
    \begin{tabular}{llccccc} %全部居中
    \toprule
    &Method & IoU $\uparrow$ &${F}_{\beta}$ $\uparrow$ & mAE $\downarrow$ &BER $\downarrow$ \\
    \midrule
    &BASNet\cite{qin2019basnet}  &69.79 &0.808 &0.106& 13.54 \\
    &MINet  \cite{pang2020multi}  &77.29 &0.879 &0.077& 9.54 \\
    &BDRAR\cite{pang2020multi} &75.92&0.860&0.081&8.61\\
    &PSPNet\cite{pang2020multi}&7.30&0.834 &0.110&10.66\\
    &MirrorNet \cite{yang2019my} &74.20 &0.828   & 0.090 & 10.76\\
    &SINet \cite{fan2020camouflaged} &77.04 &0.875&0.077 & 9.25\\&GDNet \cite{mei2020don}  &79.01 &0.869&0.069 &7.72  \\
    &TransLab\cite{xie2020segmenting} &74.05 & 0.837  &0.088  & 11.35\\
   & GlassNet \cite{lin2021rich}    &83.64 & 0.903 &0.055    & 6.12 \\
   & GlassSemNet\cite{linexploiting} &85.60 &0.920 &0.044 &5.60\\
    \midrule
   & \textbf{SAM} &\textbf{50.60} & \textbf{0.799} &\textbf{0.213} & \textbf{23.91}\\
    \bottomrule
    \end{tabular}
   %#\vspace*{-2mm}
    \caption{Experimental Comparison on GSD. %Training hyperparameters are kept consistent with the GSD-Net.}
    }
    \vspace*{-2mm}
    \label{GSD}
\end{table}

%%%%%%%%%%%%%%%%%%%%%%%%%   image

\subsection{Experiment on mirror objects}
We compare SAM with the state-of-the-art semantic segmentation, shadow detection, saliency objects detection, and mirror objects segmentation methods such as BDRAR, MirrorNet, EBLNet, PMD-Net, LSA as shown in Table \ref{MSD}, \ref{PMD}. 

Segmenting mirror objects is a challenging task in semantic segmentation due to their strong reflective properties, which cause them to have a variable appearance. To evaluate the performance of SAM model, we conducted tests on the MSD and PMD benchmark datasets, \citep{lin2020progressive}.

\begin{figure*}
\centering
\includegraphics[scale=0.8]{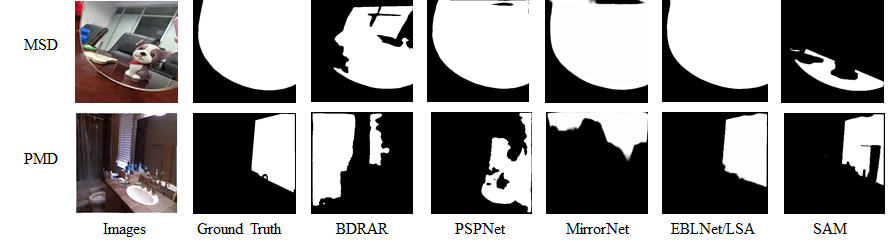}
    %\vspace*{-2mm}
\caption{Sample qualitative comparison results on MSD and PMD.}
  \label{mirror1} 
\end{figure*}

Through our experiments, we found that the PMD benchmark, with more images of mirrors captured from a distance, offers a clearer view of the boundaries of mirror regions compared to the MSD benchmark, which has more images captured from close range. As shown in Table \ref{PMD}, SAM performs comparably to state-of-the-art methods on the PMD benchmark benchmark. However, as shown in  Table\ref{MSD} and Figure\ref{mirror1}, SAM's performance on the MSD benchmark is somewhat unsatisfactory, as it tends to segment objects inside the mirror  rather than recognizing the mirror itself.

%%%%%%%%%%%%%%%%%%%%%%%%%%%%%%%%%%%%%%%%%%%%%%%%%%%%%%%%%%%%%%%%%%%%%%%%%%%%%%
\begin{table}%\scriptsize
    \centering
    \begin{tabular}{lccccc} %全部居中
    \toprule
    Method & IoU $\uparrow$ & Acc $\uparrow$ &${F}_{\beta}$ $\uparrow$ & mAE $\downarrow$ &BER $\downarrow$ \\
    \midrule
    ICNet \cite{zhao2018icnet}  & 57.25 &0.694&0.710 & 0.124 &18.75\\
    DSS\cite{hou2017deeply}  & 59.11 & 0.665  &0.743  & 0.125 & 18.81\\
    RAS\cite{chen2018reverse}  & 60.48 & 0.845 &0.758     & 0.111 & 17.60\\
    BDRAR\cite{zhu2018bidirectional}  & 67.43 & 0.821 &0.792     & 0.093 & 12.41\\
    DSC\cite{hu2018direction}  & 69.71 & 0.816  &0.812    & 0.087 & 11.77\\
    PSPNet  \cite{zhao2017pyramid} &68.01 &0.922  &0.846  &0.079 &12.08  \\
    %\midrule   
    MirrorNet\cite{yang2019my}   &78.95 & 0.935 &0.857 & 0.065 & 6.39\\
    LSA \cite{guan2022learning}  &79.85 & 0.946 & 0.889    & 0.055 & 7.12\\
    EBLNet \cite{he2021enhanced}  & 80.33  & 0.951 &0.883    & 0.049 &8.63\\
    \midrule
    \textbf{SAM}  & \textbf{51.57}&\textbf{0.876} &\textbf{0.817}& \textbf{0.124} & \textbf{23.17}\\
    \bottomrule
    \end{tabular}
        %\vspace*{-2mm}
    \caption{Experimental Comparison on MSD}
        %    \vspace*{-1mm}
        \label{MSD}

\end{table}
\begin{table}%\scriptsize
    \centering
    \begin{tabular}{lccccc} %全部居中
    \toprule
    Method & IoU $\uparrow$ &${F}_{\beta}$ $\uparrow$ & mAE $\downarrow$ &Acc $\uparrow$ \\
    \midrule
    CPNET \cite{yu2020context}  &56.36 &0.734 &0.051& 94.85 \\
    GloRe \cite{chen2019graph_global} &61.25 &0.774 &0.044 & 95.61\\
    BDRAR\cite{zhu2018bidirectional}  &  58.43 & 0.7433  & 0.043  &95.66 \\
    PSPNet  \cite{zhao2017pyramid} &60.44 &0.806    &0.039 &96.13  \\
    %\midrule
    MirrorNet \cite{yang2019my} &62.50 & 0.778  &0.041  & 96.27\\
    PMD-Net \cite{lin2020progressive}  &62.40 & 0.827 &0.055    & 96.80 \\
    LSA  \cite{guan2022learning} &66.84 & 0.844 &0.049  & 96.82\\
    \midrule
    \textbf{SAM} & \textbf{64.75} & \textbf{0.861} &\textbf{0.0525} & \textbf{94.75}\\
    \bottomrule
    \end{tabular}
        %\vspace*{-2mm}
    \caption{Experimental Comparison on PMD
    }    %\vspace*{-4mm}
    \label{PMD}
\end{table}

\subsection{Additional visualization results}

We conducted additional visualization experiments on the glass and mirror object benchmarks to facilitate observation of the results.

In contrast to the previous visualizations, we present the detection results of all objects in the predicted images of glass and mirror objects. To enhance the visibility of the results obtained from the demonstration, we applied a pseudo-binary technique to improve the clarity of the output results. Our output images are generated as pseudo-binary images, where the predicted regions of objects are represented in shades of gray, and the non-object regions are displayed as black. Furthermore, if overlapping objects are detected, the corresponding overlapping regions will be entirely displayed as white.

\begin{figure}
\centering
\includegraphics[scale=0.6]{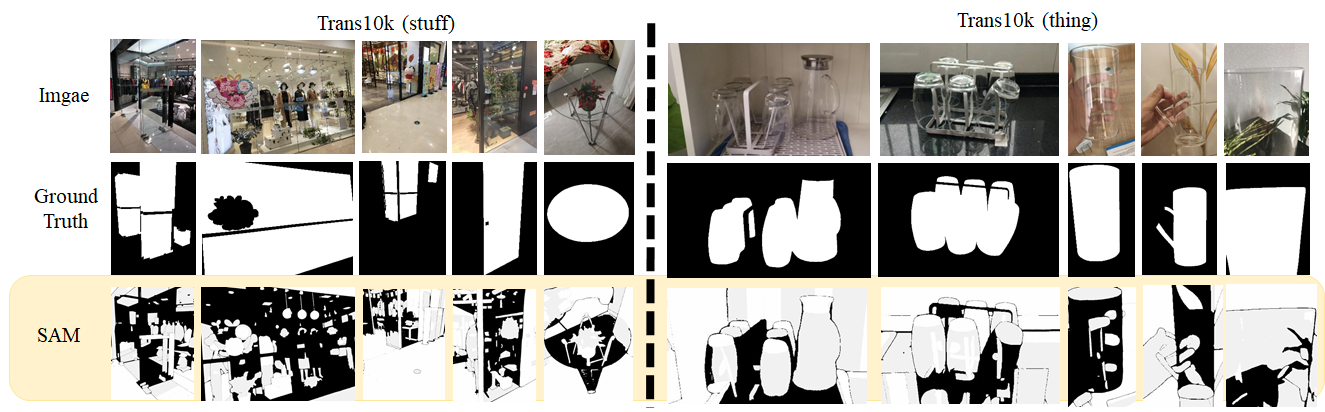}
    %\vspace*{-2mm}
\caption{Some challenging scenarios involving objects categorized as glass "things" and "stuff" in the Trans10k test sets resulted in poor testing results. Please zoom in to see the details.}
  \label{tarans1-k} 
\end{figure}

\begin{figure}
\centering
\includegraphics[scale=0.75]{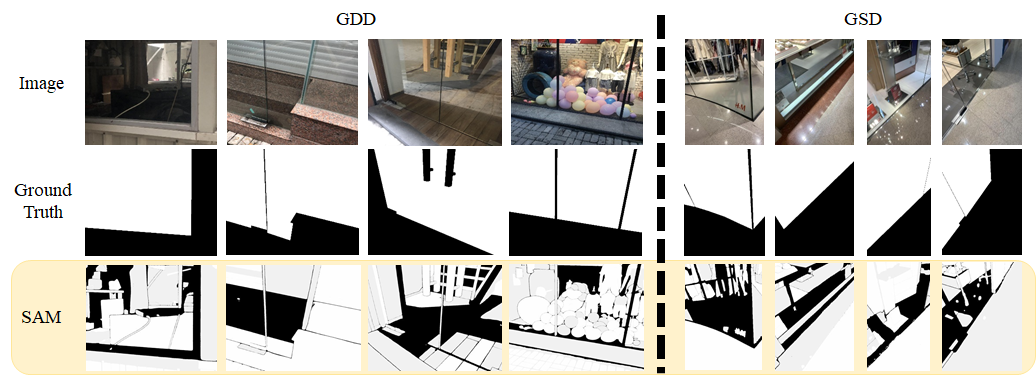}
    %\vspace*{-2mm}
\caption{Some bad testing results obtain from challenging scenarios that involve glass objects.}
  \label{glass-all} 
\end{figure}

\begin{figure}
\centering
\includegraphics[scale=0.75]{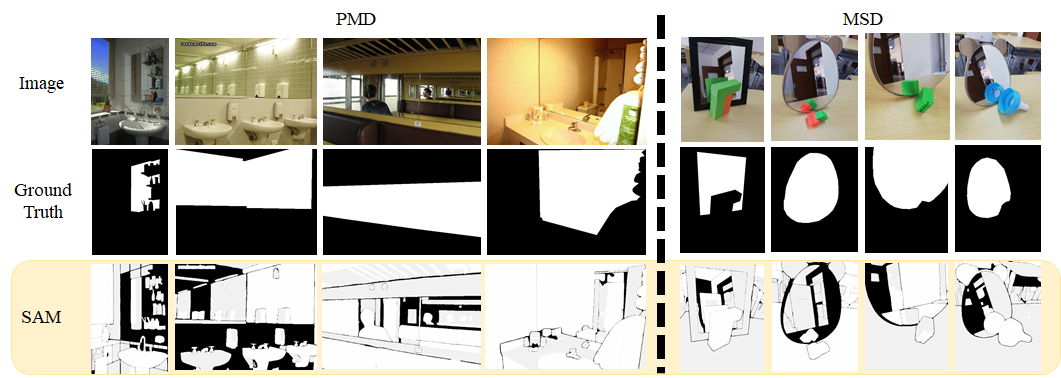}
    %\vspace*{-2mm}
\caption{Some bad testing results obtain from challenging scenarios that involve mirror objects.}
  \label{mirror-all} 
\end{figure}

As shown in Figures \ref{tarans1-k} and \ref{glass-all}, when recognizing glass objects, the SAM model successfully identified objects behind the transparent ones, but failed to recognize the glass objects themselves. And many regions inside the glass were not segmented as any object, resulting in these regions being unlabeled and displayed as black areas. In Figure \ref{mirror-all} the model correctly identified most objects reflected in the mirror, but neglected to recognize the mirror itself.

%%%%%%%%%%%%%%%%%%%%%%%%%%%%%%%%%%%%%%%%%%%%%%%
%%%%%%%%%%%%%%%%%%%%%%%%%%%%%%%%%%%%%%%%%%%%%%%%%%%%%%%%%%%%%%%%%%
% \begin{figure*}
% \centering
% \includegraphics[scale=0.7]{thing.jpg}
%     %\vspace*{-2mm}
% \caption{Some challenging scenarios involving objects categorized as glass "things" in the Trans10k test sets resulted in poor testing results.}
%   \label{thing} 
% \centering
% \includegraphics[scale=0.75]{figs_glass/stuff.jpg}
%     %\vspace*{-2mm}
% \caption{Some challenging scenarios involving objects categorized as glass "stuff" in the Trans10k test sets resulted in poor testing results.}
%   \label{stuff} 
% \end{figure*}
%%%%%%%%%%%%%%%%%%%%%%%%%%%%%%%%%%%%%%%%%%%%%%%%%%%%%%%%%%%

\section{Conclusion}
This work is the first of its kind to perform a comprehensive study on whether SAM can segment glass-related objects. We find that SAM often fails to detect transparent objects. On two benchmark datasets, the performance of SAM is significantly worse than those models that are specifically trained to detect transparent objects. 

% In this article, we conducted tests on the robustness of SAM for identifying and segmenting transparent and mirror objects. Our findings suggest that SAM, as a foundation model in the field of computer vision, holds great potential for accurately identifying and segmenting glass objects.

\bibliographystyle{unsrtnat}
\bibliography{bib_local,bib_sam}

\begin{thebibliography}{32}
\providecommand{\natexlab}[1]{#1}
\providecommand{\url}[1]{\texttt{#1}}
\expandafter\ifx\csname urlstyle\endcsname\relax
  \providecommand{\doi}[1]{doi: #1}\else
  \providecommand{\doi}{doi: \begingroup \urlstyle{rm}\Url}\fi

\bibitem[Zhang et~al.(2023{\natexlab{a}})Zhang, Zhang, Zheng, Qiao, Li, Zhang,
  Dam, Thwal, Tun, Huy, et~al.]{zhang2023complete}
Chaoning Zhang, Chenshuang Zhang, Sheng Zheng, Yu~Qiao, Chenghao Li, Mengchun
  Zhang, Sumit~Kumar Dam, Chu~Myaet Thwal, Ye~Lin Tun, Le~Luang Huy, et~al.
\newblock A complete survey on generative ai (aigc): Is chatgpt from gpt-4 to
  gpt-5 all you need?
\newblock \emph{arXiv preprint arXiv:2303.11717}, 2023{\natexlab{a}}.

\bibitem[Zhang et~al.(2023{\natexlab{b}})Zhang, Zhang, Li, Qiao, Zheng, Dam,
  Zhang, Kim, Kim, Choi, et~al.]{zhang2023chatgpt}
Chaoning Zhang, Chenshuang Zhang, Chenghao Li, Yu~Qiao, Sheng Zheng,
  Sumit~Kumar Dam, Mengchun Zhang, Jung~Uk Kim, Seong~Tae Kim, Jinwoo Choi,
  et~al.
\newblock One small step for generative ai, one giant leap for agi: A complete
  survey on chatgpt in aigc era.
\newblock \emph{arXiv preprint arXiv:2304.06488}, 2023{\natexlab{b}}.

\bibitem[Zhang et~al.(2023{\natexlab{c}})Zhang, Zhang, Zhang, and
  Kweon]{zhang2023text}
Chenshuang Zhang, Chaoning Zhang, Mengchun Zhang, and In~So Kweon.
\newblock Text-to-image diffusion models in generative ai: A survey.
\newblock \emph{arXiv preprint arXiv:2303.07909}, 2023{\natexlab{c}}.

\bibitem[Zhang et~al.(2023{\natexlab{d}})Zhang, Zhang, Zheng, Zhang, Qamar,
  Bae, and Kweon]{zhang2023audio}
Chenshuang Zhang, Chaoning Zhang, Sheng Zheng, Mengchun Zhang, Maryam Qamar,
  Sung-Ho Bae, and In~So Kweon.
\newblock A survey on audio diffusion models: Text to speech synthesis and
  enhancement in generative ai.
\newblock \emph{arXiv preprint arXiv:2303.13336}, 2023{\natexlab{d}}.

\bibitem[Zhang et~al.(2023{\natexlab{e}})Zhang, Qamar, Kang, Jung, Zhang, Bae,
  and Zhang]{zhang2023graph_survey}
Mengchun Zhang, Maryam Qamar, Taegoo Kang, Yuna Jung, Chenshuang Zhang, Sung-Ho
  Bae, and Chaoning Zhang.
\newblock A survey on graph diffusion models: Generative ai in science for
  molecule, protein and material.
\newblock \emph{arXiv preprint arXiv:2304.01565}, 2023{\natexlab{e}}.

\bibitem[Bommasani et~al.(2021)Bommasani, Hudson, Adeli, Altman, Arora, von
  Arx, Bernstein, Bohg, Bosselut, Brunskill,
  et~al.]{bommasani2021opportunities}
Rishi Bommasani, Drew~A Hudson, Ehsan Adeli, Russ Altman, Simran Arora, Sydney
  von Arx, Michael~S Bernstein, Jeannette Bohg, Antoine Bosselut, Emma
  Brunskill, et~al.
\newblock On the opportunities and risks of foundation models.
\newblock \emph{arXiv preprint arXiv:2108.07258}, 2021.

\bibitem[Brown et~al.(2020)Brown, Mann, Ryder, Subbiah, Kaplan, Dhariwal,
  Neelakantan, Shyam, Sastry, Askell, et~al.]{brown2020language}
Tom Brown, Benjamin Mann, Nick Ryder, Melanie Subbiah, Jared~D Kaplan, Prafulla
  Dhariwal, Arvind Neelakantan, Pranav Shyam, Girish Sastry, Amanda Askell,
  et~al.
\newblock Language models are few-shot learners.
\newblock \emph{Advances in neural information processing systems},
  33:\penalty0 1877--1901, 2020.

\bibitem[Kirillov et~al.(2023)Kirillov, Mintun, Ravi, Mao, Rolland, Gustafson,
  Xiao, Whitehead, Berg, Lo, et~al.]{kirillov2023segment}
Alexander Kirillov, Eric Mintun, Nikhila Ravi, Hanzi Mao, Chloe Rolland, Laura
  Gustafson, Tete Xiao, Spencer Whitehead, Alexander~C Berg, Wan-Yen Lo, et~al.
\newblock Segment anything.
\newblock \emph{arXiv preprint arXiv:2304.02643}, 2023.

\bibitem[Lin et~al.(2021)Lin, He, and Lau]{lin2021rich}
Jiaying Lin, Zebang He, and Rynson~WH Lau.
\newblock Rich context aggregation with reflection prior for glass surface
  detection.
\newblock In \emph{Proceedings of the IEEE/CVF Conference on Computer Vision
  and Pattern Recognition}, pages 13415--13424, 2021.

\bibitem[Maeno et~al.(2013)Maeno, Nagahara, Shimada, and
  Taniguchi]{maeno2013light}
Kazuki Maeno, Hajime Nagahara, Atsushi Shimada, and Rin-ichiro Taniguchi.
\newblock Light field distortion feature for transparent object recognition.
\newblock In \emph{Proceedings of the IEEE Conference on Computer Vision and
  Pattern Recognition}, pages 2786--2793, 2013.

\bibitem[He et~al.(2021)He, Li, Cheng, Shi, Tong, Meng, Prinet, and
  Weng]{he2021enhanced}
Hao He, Xiangtai Li, Guangliang Cheng, Jianping Shi, Yunhai Tong, Gaofeng Meng,
  V{\'e}ronique Prinet, and LuBin Weng.
\newblock Enhanced boundary learning for glass-like object segmentation.
\newblock In \emph{Proceedings of the IEEE/CVF International Conference on
  Computer Vision}, pages 15859--15868, 2021.

\bibitem[Mei et~al.(2020)Mei, Yang, Wang, Liu, He, Zhang, Wei, and
  Lau]{mei2020don}
Haiyang Mei, Xin Yang, Yang Wang, Yuanyuan Liu, Shengfeng He, Qiang Zhang,
  Xiaopeng Wei, and Rynson~WH Lau.
\newblock Don't hit me! glass detection in real-world scenes.
\newblock In \emph{Proceedings of the IEEE/CVF Conference on Computer Vision
  and Pattern Recognition}, pages 3687--3696, 2020.

\bibitem[Yang et~al.(2019)Yang, Mei, Xu, Wei, Yin, and Lau]{yang2019my}
Xin Yang, Haiyang Mei, Ke~Xu, Xiaopeng Wei, Baocai Yin, and Rynson~WH Lau.
\newblock Where is my mirror?
\newblock In \emph{Proceedings of the IEEE/CVF International Conference on
  Computer Vision}, pages 8809--8818, 2019.

\bibitem[Lin et~al.(2020)Lin, Wang, and Lau]{lin2020progressive}
Jiaying Lin, Guodong Wang, and Rynson~WH Lau.
\newblock Progressive mirror detection.
\newblock In \emph{Proceedings of the IEEE/CVF Conference on Computer Vision
  and Pattern Recognition}, pages 3697--3705, 2020.

\bibitem[Xie et~al.(2020)Xie, Wang, Wang, Ding, Shen, and
  Luo]{xie2020segmenting}
Enze Xie, Wenjia Wang, Wenhai Wang, Mingyu Ding, Chunhua Shen, and Ping Luo.
\newblock Segmenting transparent objects in the wild.
\newblock In \emph{Computer Vision--ECCV 2020: 16th European Conference,
  Glasgow, UK, August 23--28, 2020, Proceedings, Part XIII 16}, pages 696--711.
  Springer, 2020.

\bibitem[Margolin et~al.(2014)Margolin, Zelnik-Manor, and
  Tal]{margolin2014evaluate}
Ran Margolin, Lihi Zelnik-Manor, and Ayellet Tal.
\newblock How to evaluate foreground maps?
\newblock In \emph{Proceedings of the IEEE conference on computer vision and
  pattern recognition}, pages 248--255, 2014.

\bibitem[Achanta et~al.(2009)Achanta, Hemami, Estrada, and
  Susstrunk]{achanta2009frequency}
Radhakrishna Achanta, Sheila Hemami, Francisco Estrada, and Sabine Susstrunk.
\newblock Frequency-tuned salient region detection.
\newblock In \emph{2009 IEEE conference on computer vision and pattern
  recognition}, pages 1597--1604. IEEE, 2009.

\bibitem[Zhao et~al.(2017)Zhao, Shi, Qi, Wang, and Jia]{zhao2017pyramid}
Hengshuang Zhao, Jianping Shi, Xiaojuan Qi, Xiaogang Wang, and Jiaya Jia.
\newblock Pyramid scene parsing network.
\newblock In \emph{Proceedings of the IEEE conference on computer vision and
  pattern recognition}, pages 2881--2890, 2017.

\bibitem[Kirillov et~al.(2020)Kirillov, Wu, He, and
  Girshick]{kirillov2020pointrend}
Alexander Kirillov, Yuxin Wu, Kaiming He, and Ross Girshick.
\newblock Pointrend: Image segmentation as rendering.
\newblock In \emph{Proceedings of the IEEE/CVF conference on computer vision
  and pattern recognition}, pages 9799--9808, 2020.

\bibitem[Liu et~al.(2018)Liu, Han, and Yang]{liu2018picanet}
Nian Liu, Junwei Han, and Ming-Hsuan Yang.
\newblock Picanet: Learning pixel-wise contextual attention for saliency
  detection.
\newblock In \emph{Proceedings of the IEEE conference on computer vision and
  pattern recognition}, pages 3089--3098, 2018.

\bibitem[Hu et~al.(2018)Hu, Zhu, Fu, Qin, and Heng]{hu2018direction}
Xiaowei Hu, Lei Zhu, Chi-Wing Fu, Jing Qin, and Pheng-Ann Heng.
\newblock Direction-aware spatial context features for shadow detection.
\newblock In \emph{Proceedings of the IEEE conference on computer vision and
  pattern recognition}, pages 7454--7462, 2018.

\bibitem[Zhu et~al.(2018)Zhu, Deng, Hu, Fu, Xu, Qin, and
  Heng]{zhu2018bidirectional}
Lei Zhu, Zijun Deng, Xiaowei Hu, Chi-Wing Fu, Xuemiao Xu, Jing Qin, and
  Pheng-Ann Heng.
\newblock Bidirectional feature pyramid network with recurrent attention
  residual modules for shadow detection.
\newblock In \emph{Proceedings of the European Conference on Computer Vision
  (ECCV)}, pages 121--136, 2018.

\bibitem[Qin et~al.(2019)Qin, Zhang, Huang, Gao, Dehghan, and
  Jagersand]{qin2019basnet}
Xuebin Qin, Zichen Zhang, Chenyang Huang, Chao Gao, Masood Dehghan, and Martin
  Jagersand.
\newblock Basnet: Boundary-aware salient object detection.
\newblock In \emph{Proceedings of the IEEE/CVF conference on computer vision
  and pattern recognition}, pages 7479--7489, 2019.

\bibitem[Pang et~al.(2020)Pang, Zhao, Zhang, and Lu]{pang2020multi}
Youwei Pang, Xiaoqi Zhao, Lihe Zhang, and Huchuan Lu.
\newblock Multi-scale interactive network for salient object detection.
\newblock In \emph{Proceedings of the IEEE/CVF conference on computer vision
  and pattern recognition}, pages 9413--9422, 2020.

\bibitem[Fan et~al.(2020)Fan, Ji, Sun, Cheng, Shen, and
  Shao]{fan2020camouflaged}
Deng-Ping Fan, Ge-Peng Ji, Guolei Sun, Ming-Ming Cheng, Jianbing Shen, and Ling
  Shao.
\newblock Camouflaged object detection.
\newblock In \emph{Proceedings of the IEEE/CVF conference on computer vision
  and pattern recognition}, pages 2777--2787, 2020.

\bibitem[Lin et~al.()Lin, Yeung, and Lau]{linexploiting}
Jiaying Lin, Yuen~Hei Yeung, and Rynson~WH Lau.
\newblock Exploiting semantic relations for glass surface detection.
\newblock In \emph{Advances in Neural Information Processing Systems}.

\bibitem[Zhao et~al.(2018)Zhao, Qi, Shen, Shi, and Jia]{zhao2018icnet}
Hengshuang Zhao, Xiaojuan Qi, Xiaoyong Shen, Jianping Shi, and Jiaya Jia.
\newblock Icnet for real-time semantic segmentation on high-resolution images.
\newblock In \emph{Proceedings of the European conference on computer vision
  (ECCV)}, pages 405--420, 2018.

\bibitem[Hou et~al.(2017)Hou, Cheng, Hu, Borji, Tu, and Torr]{hou2017deeply}
Qibin Hou, Ming-Ming Cheng, Xiaowei Hu, Ali Borji, Zhuowen Tu, and Philip~HS
  Torr.
\newblock Deeply supervised salient object detection with short connections.
\newblock In \emph{Proceedings of the IEEE conference on computer vision and
  pattern recognition}, pages 3203--3212, 2017.

\bibitem[Chen et~al.(2018)Chen, Tan, Wang, and Hu]{chen2018reverse}
Shuhan Chen, Xiuli Tan, Ben Wang, and Xuelong Hu.
\newblock Reverse attention for salient object detection.
\newblock In \emph{Proceedings of the European conference on computer vision
  (ECCV)}, pages 234--250, 2018.

\bibitem[Guan et~al.(2022)Guan, Lin, and Lau]{guan2022learning}
Huankang Guan, Jiaying Lin, and Rynson~WH Lau.
\newblock Learning semantic associations for mirror detection.
\newblock In \emph{Proceedings of the IEEE/CVF Conference on Computer Vision
  and Pattern Recognition}, pages 5941--5950, 2022.

\bibitem[Yu et~al.(2020)Yu, Wang, Gao, Yu, Shen, and Sang]{yu2020context}
Changqian Yu, Jingbo Wang, Changxin Gao, Gang Yu, Chunhua Shen, and Nong Sang.
\newblock Context prior for scene segmentation.
\newblock In \emph{Proceedings of the IEEE/CVF conference on computer vision
  and pattern recognition}, pages 12416--12425, 2020.

\bibitem[Chen et~al.(2019)Chen, Rohrbach, Yan, Shuicheng, Feng, and
  Kalantidis]{chen2019graph_global}
Yunpeng Chen, Marcus Rohrbach, Zhicheng Yan, Yan Shuicheng, Jiashi Feng, and
  Yannis Kalantidis.
\newblock Graph-based global reasoning networks.
\newblock In \emph{CVPR}, pages 433--442, 2019.

\end{thebibliography}
%\bibliography{bib_sam}
\end{document}